\documentclass[11pt]{article}
\usepackage{amsfonts}
\usepackage{amssymb}
\usepackage{amstext}
\usepackage{amsmath}
\usepackage{xspace}
\usepackage{theorem}
\usepackage{graphicx}
\usepackage{url}
\usepackage{graphics}
\usepackage{colordvi}
\usepackage{colordvi}
\usepackage{subfig}
\usepackage{caption}
\usepackage{hyperref}
\begin{document}

\title{A Multiscale Patch Based Convolutional Network for Brain Tumor Segmentation. }

\author{Jean Stawiaski\\ \small Stryker Corporation, Freiburg, Germany.}

\date{September 22, 2017.}

% -----------------------------------------------------------------------------
\maketitle

\begin{abstract}
  This article presents a multiscale patch based convolutional neural network for the automatic segmentation of brain tumors in multi-modality 3D MR images. We use multiscale deep supervision and inputs to train a convolutional network. We evaluate the effectiveness of the proposed approach on the BRATS 2017 segmentation challenge \cite{brats1,brats2,brats3,brats4} where we obtained dice scores of 0.755, 0.900, 0.782 and 95\% Hausdorff distance of 3.63mm, 4.10mm, and 6.81mm  for enhanced tumor core, whole tumor and tumor core respectively.
\end{abstract}


% -----------------------------------------------------------------------------
\section{Introduction}

Brain Gliomas represent 80\% of all malignant brain tumors. Gliomas can be categorized according to their grade which is determined by a pathologic evaluation of the tumor:
\begin{itemize}
\item Low-grade gliomas exhibit benign tendencies and indicate thus a better prognosis for the patient. However, they also have a uniform rate of recurrence and increase in grade over time.
\item High-grade gliomas are anaplastic; these are malignant and carry a worse prognosis for the patient.
\end{itemize}

\begin{figure}[httb]
    \centering
    \includegraphics[scale=0.37]{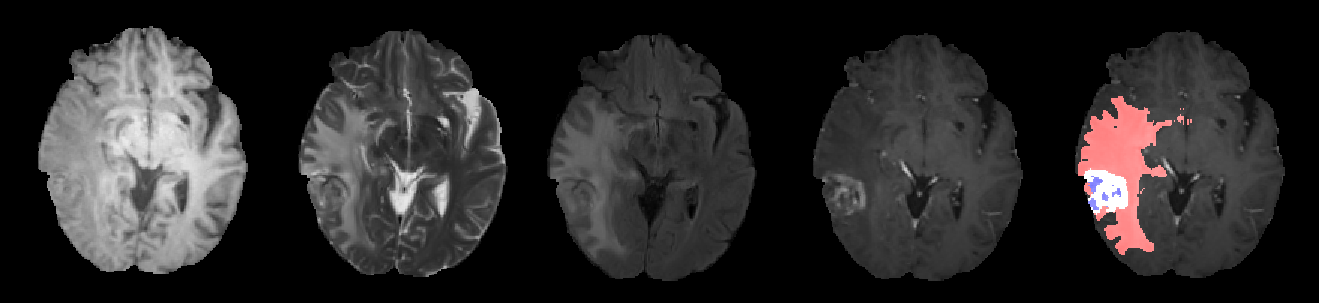}
    \caption{Example of images from the BRATS 2017 dataset. From left to right: T1 image, T2 image: the whole tumor and its core are visible, T2 FLAIR image: discarding the cerebrospinal fluid signal from the T2 image highlights the tumor region only, T1ce: contrast injection permits to visualize the enhancing part of the tumor as well as the necrotic part. Finally the expected segmentation result is overlaid on the T1ce image. The edema is shown in red, the enhancing part in white and the necrotic part of the tumor is shown in blue. }
    \label{bratsex}
\end{figure}

Brain gliomas can be well detected using magnetic resonance imaging. The whole tumor is visible in T2-FLAIR, the tumor core is visible in T2 and the enhancing tumor structures as well as the necrotic parts can be visualized using contrast enhanced T1 scans. An example is illustrated in figure \ref{bratsex}. \\

Automatic segmentation of brain tumor structures is particularly important in order to quantitatively assess the tumor geometry. It has also a great potential for surgical planning and intraoperative surgical resection guidance. Automatic segmentation still poses many challenges because of the variability of appearances and sizes of the tumors. Moreover the differences in the image acquisition protocols, the inhomogeneity of the magnetic field and partial volume effects have also a great impact on the image quality obtained from routinely acquired 3D MR images.\\

In the recent years, deep neural networks have shown to provide state-of-the-art performance for various challenging image segmentation and classification problems \cite{FCN,FCN-CRF,segnet,dilnet,Deconv}. Medical image segmentation problems have also been successfully tackled by such approaches \cite{UNET,VNET,deepsuper2,deepsuper3,cascade}. Inspired by these works, we present here a relatively simple architecture that produces competitive results for the BRATS 2017 dataset \cite{brats1,brats2,brats3,brats4}. We propose a variant of the well known U-net \cite{UNET}, fed with multiscale inputs \cite{multiscalenet}, having residual connections \cite{ResNet}, and being trained in a multiscale deep supervised manner \cite{deepsuper}.\\

% -----------------------------------------------------------------------------
\section{Multiscale Patch Based Convolutional Network}

This section details our network architecture, the loss function used to train the network end-to-end as well as the training data preparation.

% -----------------------------------------------------------------------------
\subsection{Network Architecture}

\begin{figure}[httb]
    \centering
    \includegraphics[scale=0.3]{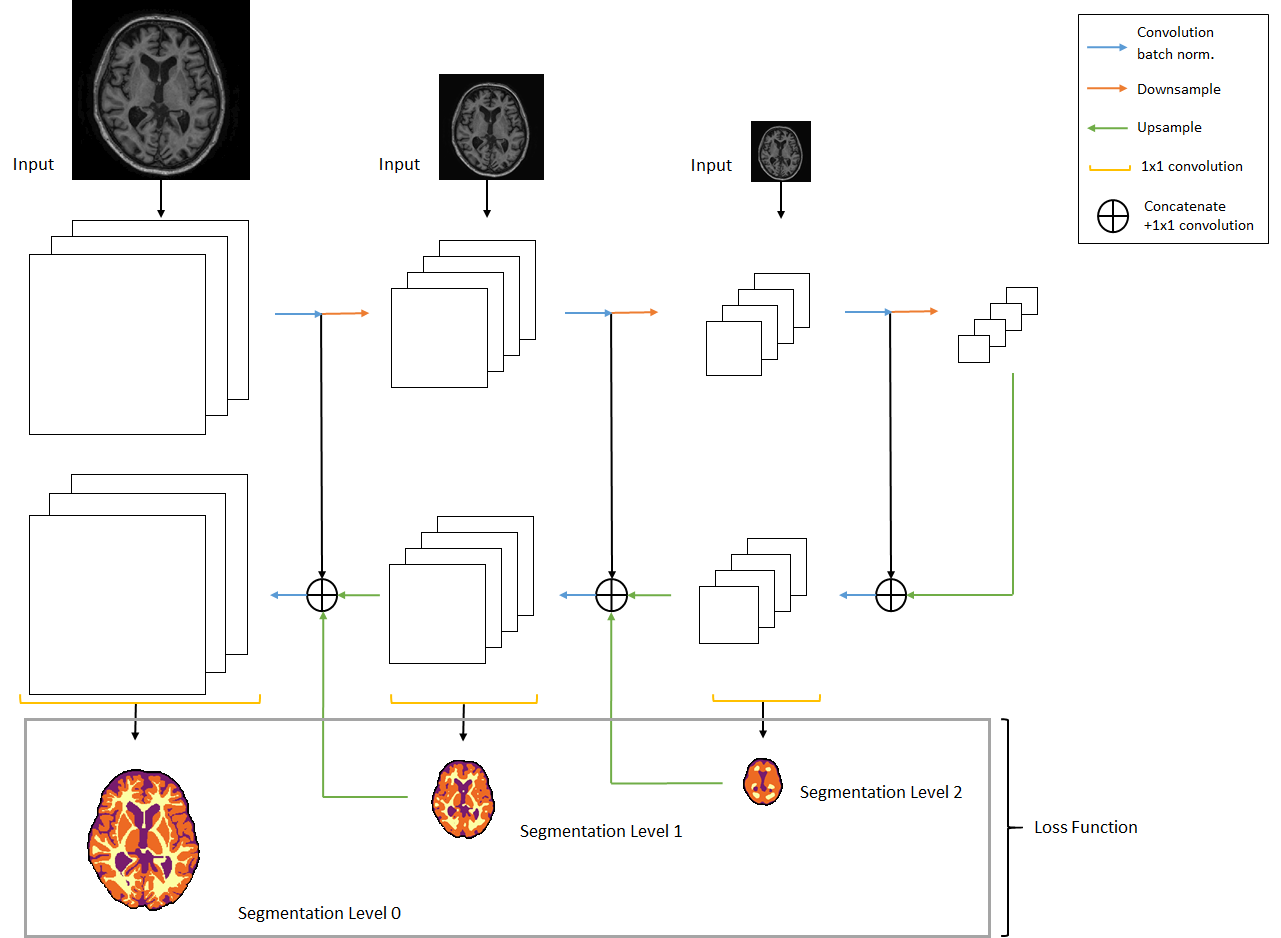}
    \caption{Network architecture: multiscale convolutional neural network.}
    \label{archi}
\end{figure}

\begin{figure}[httb]
    \centering
    \includegraphics[scale=0.2]{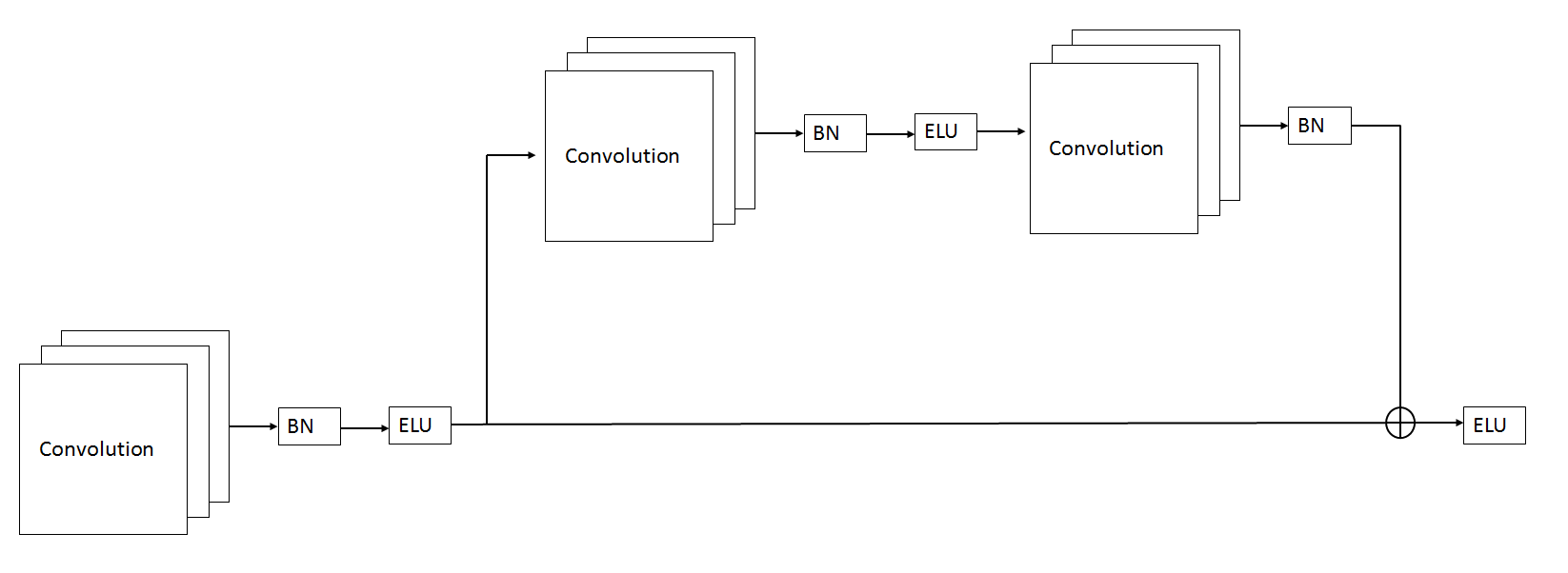}
    \caption{Residual connections in a convolutional layer.}
    \label{residual}
\end{figure}

\begin{figure}[httb]
    \centering
    \includegraphics[scale=0.33]{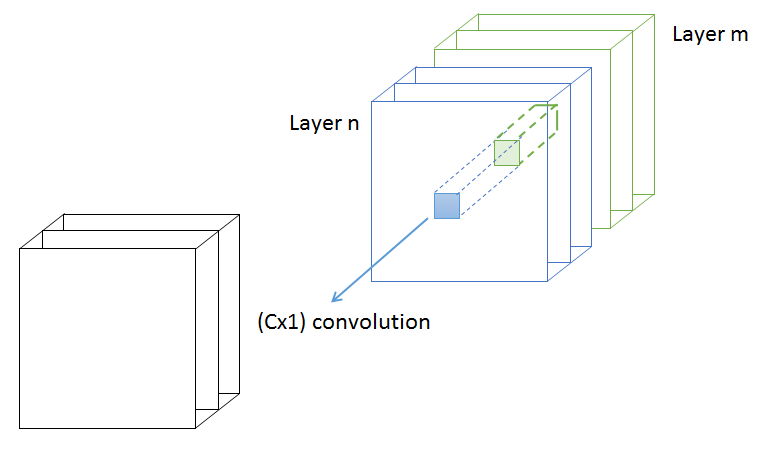}
    \caption{Layer combination using (1x1) convolution kernels.}
    \label{combination}
\end{figure}

Our architecture is illustrated in figure \ref{archi}. The network processes patches of $64^3$ voxels and takes multiscale version of these patches as input. We detail here some important properties of the network:

\begin{itemize}

\item each sample image $y$ is normalised to have zero mean and unit variance for voxels inside the brain:
\begin{equation}
y = \frac{x-m_{br} }{\sigma_{br} } \; ,
\end{equation}
where $m_{br}$ and $\sigma_{br}$ is the mean and the variance of voxels inside the brain (non zero voxels of a given 3D image),

\item batch normalisation is performed after each convolutional layer using a running mean and standard deviation computed on 5000 samples:
\begin{equation}
by = \frac{ (y-m_b) }{( \sigma_b + \epsilon) } \times \gamma + c \; ,
\end{equation}
where $m_{b}$ and $\sigma_{b}$ is the mean and variance of the minibatches and $\gamma$ and $c$ are learnable parameters,

\item each layer is composed of residual connections as illustrated in figure \ref{residual},

\item different layers of the network are combined using (1x1) convolution kernels as illustrated in figure \ref{combination},

\item the activation function is an exponential linear unit,

\item convolution kernels are (3x3x3) kernels,

\item convolutions are computed using reflective border padding,

\item downsampling is performed by decimating a smooth version of the layer:
\begin{equation}
dy = (y  \ast G_\sigma) \downarrow_2 \; ,
\end{equation}
where $G_\sigma$ is a gaussian kernel,

\item upsampling is performed by nearest neighbor interpolation.

\end{itemize}

% -----------------------------------------------------------------------------
\subsection{Scale-wise Loss Function}

We define a loss function for each scale of the network allowing a faster training and a better model convergence. Using this deep supervision, gradients are efficiently injected at all scales of the network during the training process. Downsampled ground truth segmentation images are used to compute the loss associated for each scale. The loss function is defined as the combination of the mean cross entropy (mce) and the Dice coefficients (dce) between the ground truth class probability and the network estimates:
\begin{equation}
ce = \sum_k \Big( \frac{-1}{n} \sum_i y_i^k log(p_i^k) \Big) \;
\end{equation}
where $y_i^k$ and $p_i^k$ represent respectively the ground truth probability and the network estimate for the class $k$ at location $i$.

\begin{equation}
dce = \sum_{k \neq 0}  \Big( 1.0 - \frac{1}{n} \Big( \frac{ 2 \sum_i p_i^k y_i^k }{ \sum_i (p_i^k)^2 + \sum_i (y_i^k)^2  } \Big) \Big) \; .
\end{equation}

Note that we exclude the background class for the computation of the dice coefficient.

% -----------------------------------------------------------------------------
\subsection{Training Data Preparation}

We used the BRATS 2017 training and validation sets for our experiments \cite{brats1,brats2,brats3,brats4}. The training set contains 285 patients (210 high grade gliomas and 75 low grade gliomas). The BRATS 2017 validation set contains 46 patients with brain tumors of unknown grade with unknown ground truth segmentations. Each patient contains four modalities: T1, T1 with contrast enhancement, T2 and T2 FLAIR. The aim of this experiment is to segment automatically the tumor necrotic part, the tumor edema and the tumor enhancing part.\\

\begin{figure}[h]
    \centering
    \includegraphics[scale=0.2]{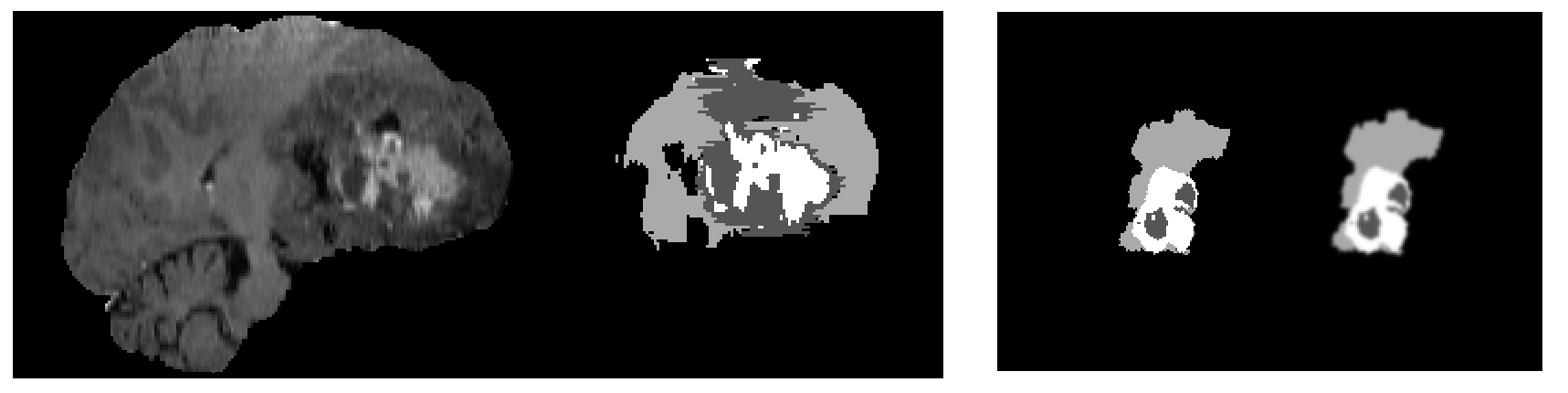}
    \caption{Noisy segmentation ground truth. Example of class wise probability smoothing. (Necrotic parts is shown in dark gray, edema in light gray and enhancing tumor in white). }
    \label{noisy}
\end{figure}

The segmentation ground truth provided with the BRATS 2017 dataset presents however some imperfections. The segmentation is relatively noisy and does not present a strong 3D coherence as illustrated in figure \ref{noisy}. We have thus decided to manually smooth each ground truth segmentation map independently such that:
\begin{equation}
y^k = (y^k \ast G_\sigma) \; ,
\end{equation}
where $y^k $ is the probability map associated with the class $k$, $G_\sigma$ is a normalised gaussian kernel. Note that this process still ensures that $y_k$ is a probability map:
\begin{equation}
\sum_{k} y_i^k = 1 \; .
\end{equation}

In order to deal with the class imbalance, patches are sampled so that at least 0.01 \% of the voxels contain one of the tumor classes.

% -----------------------------------------------------------------------------
\subsection{Implementation}

The network is implemented using Microsoft CNTK \footnote{\url{https://www.microsoft.com/en-us/cognitive-toolkit/}}. We use stochastic gradient descent with momentum to train the network. The network is trained using 3 Nvidia GTX 1080 receiving a different subset of the training data. The inference of the network is done on one graphic card and takes approximatively 5 seconds to process an image by analyzing non overlapping sub volumes of $64^3$ voxels.\\

% -----------------------------------------------------------------------------
\section{Results}

\begin{figure}[httb]
    \centering
    \includegraphics[scale=0.3]{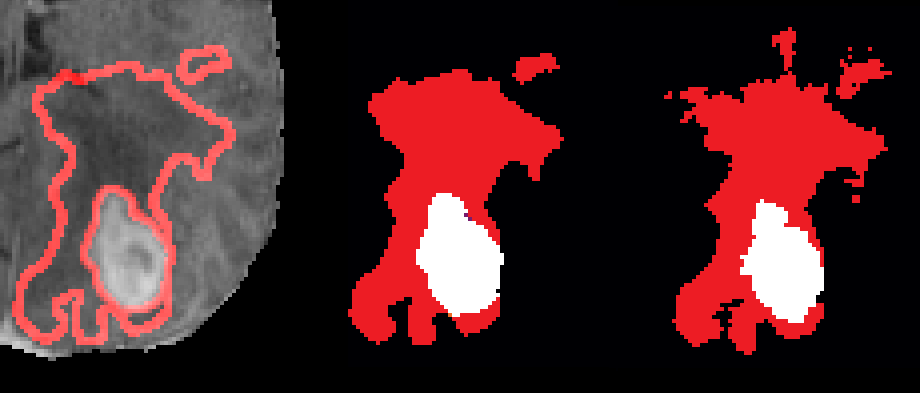}
    \caption{Left: segmentation obtained on a image from the training data. Middle: obtained segmentation result. Right: Ground truth segmentation. The edema is shown in red, the enhancing part in white and the necrotic part of the tumor is shown in blue. Our results tend to be smoother than the ground truth delineation. }
    \label{train}
\end{figure}

Due to our training data preparation (class wise segmentation smoothing) and due to our data augmentation method (additive noise), our segmentation results tends to be smoother than the ground truth segmentation. This effect is illustrated in figure \ref{train}.\\

We uploaded our segmentation results to the BRATS 2017 server \footnote{\url{https://www.cbica.upenn.edu/BraTS17/lboardValidation.html} } which evaluates the segmentation and provides quantitative measurements in terms of Dice scores, sensitivity, specificity and Hausdorff distances of enhanced tumor core, whole tumor, and tumor core. Results of the BRATS 2017 validation phase are presented in Table 1. The table summarizes the scores as they appeared on the leaderboard the 22 September 2017. We observe that the proposed method does not perform as well as the other best methods in terms of dice coefficients. On the other side our method produces very competitive distances metrics.\\

\begin{center}
  \begin{tabular}{ | l | l | l | l | l | l | l | }
    \hline
    Team & Dice ET & Dice WT & Dice TC & Dist. ET & Dist. WT & Dist. TC \\ \hline
    UCL-TIG   & 0.785 & 0.904 & 0.837 & 3.28 & 3.89 & 6.47 \\ \hline
    biomedia1 & 0.757 & 0.901 & 0.820 & 4.22 & 4.55 & 6.10 \\ \hline
    Zhouch    & 0.760 & 0.903 & 0.824 & 3.71 & 4.87 & 6.74 \\ \hline
    MIC DKFZ  & 0.731 & 0.896 & 0.797 & 4.54 & 6.97 & 9.47 \\ \hline
    \textbf{stryker}  & \textbf{0.755} & \textbf{0.900} & \textbf{0.782} & \textbf{3.63} & \textbf{4.10} & \textbf{6.81} \\
    \hline
  \end{tabular}
  \captionof{table}{BRATS 2017 Validation scores, dice coefficients and the 95\% Hausdorff distances. Our results corresponds to the team name "stryker". }
\end{center}

\begin{figure}[httb]
    \centering
    \includegraphics[scale=0.45]{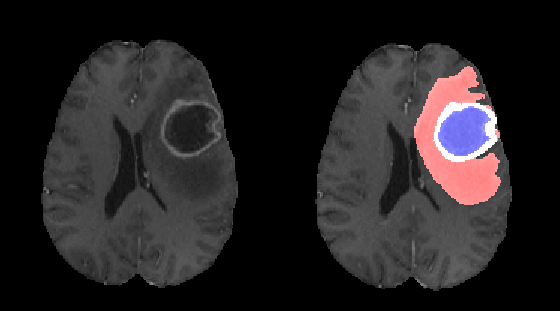}
    \includegraphics[scale=0.4]{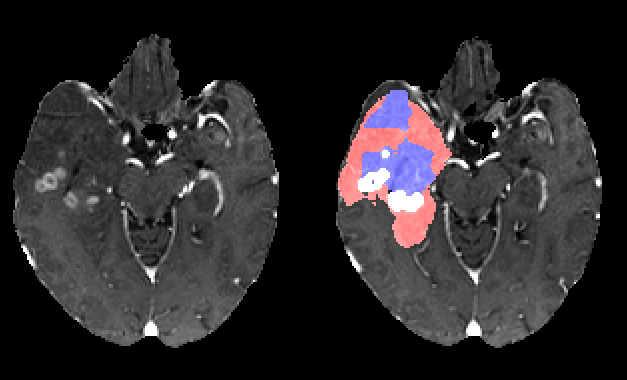}
    \caption{Segmentation results obtained on images from the validation data. (Top: good results, Bottom: incorrect detection of necrotic parts.) The edema is shown in red, the enhancing part in white and the necrotic part of the tumor is shown in blue.  }
    \label{test1}
\end{figure}

Different segmentation results are illustrated in figure \ref{test1}. The proposed network tends to produce smooth and compact segmentation results which are often very close in terms of Euclidean distance to the ground truth segmentation. We have consciously chosen to privilege this effect by smoothing the ground truth segmentation and augmenting data with additive noise. Different approaches may be better suited for other kind of quality metrics.\\

% -----------------------------------------------------------------------------
\section{Conclusion}

We have presented a relatively simple but efficient approach for automatic brain tumor segmentation using a convolutional network. We obtained competitive scores on the BRATS 2017 segmentation challenge \footnote{\url{https://www.cbica.upenn.edu/BraTS17/lboardValidation.html}}. Future work will concentrate on making the network more compact and more robust in order to be used clinically in an intraoperative setup. A possible improvement of the presented method could consist in adding semantic constraints by using a hierarchical approach such as the one presented in \cite{UCL}.\\

\bibliographystyle{amsplain}

\end{document}